\documentclass[letterpaper, 10 pt, conference]{ieeeconf}

\makeatletter
\def\endthebibliography{%
  \def\@noitemerr{\@latex@warning{Empty `thebibliography' environment}}%
  \endlist
}
\makeatother
 
\IEEEoverridecommandlockouts
\usepackage{cite}
\usepackage{amsmath,amssymb,amsfonts}
\usepackage{graphicx}
\usepackage{textcomp}
\usepackage{xcolor}
\usepackage{todonotes}
\usepackage{amsmath}
\usepackage{leftidx}
\usepackage{siunitx}
\usepackage{multirow}
\usepackage{footnote}
\usepackage[utf8]{inputenc}
\usepackage{array}
\usepackage{mdwmath}
\usepackage{mdwtab}
\usepackage{eqparbox}
\usepackage{url}
\usepackage{cite}
\usepackage{amsmath,amssymb,amsfonts}
\usepackage{algorithm2e}
\RestyleAlgo{ruled}
\usepackage{textcomp}
\usepackage{lettrine}
\usepackage{caption}
\usepackage{subcaption}
\usepackage[export]{adjustbox}
\usepackage{amsmath}
\usepackage{leftidx}
\usepackage{siunitx}
\usepackage{multirow}
\usepackage{balance}
\usepackage[export]{adjustbox}
\usepackage{booktabs}
\usepackage{hyperref}
\usepackage{balance}

\usepackage{makecell}

\renewcommand{\arraystretch}{1.00}

\newcommand{\boldparagraph}[1]{\vspace{0.3em}\noindent{\bf #1} }

\def\BibTeX{{\rm B\kern-.05em{\sc i\kern-.025em b}\kern-.08em
    T\kern-.1667em\lower.7ex\hbox{E}\kern-.125emX}}

\begin{document}





\title{\LARGE \bf Learning High-level Semantic-Relational Concepts for SLAM}



\author{Jose Andres Millan-Romera$^{1}$, Hriday Bavle$^{1}$, Muhammad Shaheer$^{1}$, \\ Martin R. Oswald$^{2}$, Holger Voos$^{1}$, and Jose Luis Sanchez-Lopez$^{1}$ 
\thanks{$^{1}$Authors are with the Automation and Robotics Research Group, Interdisciplinary Centre for Security, Reliability and Trust (SnT), University of Luxembourg. Holger Voos is also associated with the Faculty of Science, Technology and Medicine, University of Luxembourg, Luxembourg.
\tt{\small{\{jose.millan, hriday.bavle, muhammad.shaheer, holger.voos, joseluis.sanchezlopez\}}@uni.lu}}%
\thanks{$^{2}$Author is with the University of Amsterdam.
\tt{\small{m.r.oswald@uva.nl}}}%
\thanks{*
This work was partially funded by the Fonds National de la Recherche of Luxembourg (FNR) under the projects 17097684/RoboSAUR and C22/IS/17387634/DEUS.}%
\thanks{*
For the purpose of Open Access, and in fulfillment of the obligations arising from the grant agreement, the authors have applied a Creative Commons Attribution 4.0 International (CC BY 4.0) license to any Author Accepted Manuscript version arising from this submission.}
}

\maketitle

\begin{abstract} \label{abstract}
Recent works on SLAM extend their pose graphs with higher-level semantic concepts like \textit{Rooms} exploiting relationships between them, to provide, not only a richer representation of the situation/environment but also to improve the accuracy of its estimation.
Concretely, our previous work, Situational Graphs (\textit{S-Graphs+}), a pioneer in jointly leveraging semantic relationships in the factor optimization process, relies on semantic entities such as \textit{Planes} and \textit{Rooms}, whose relationship is mathematically defined.
Nevertheless, there is no unique approach to finding all the hidden patterns in lower-level factor-graphs that correspond to high-level concepts of different natures. It is currently tackled with ad-hoc algorithms, which limits its graph expressiveness.

To overcome this limitation, in this work, we propose an algorithm based on Graph Neural Networks for learning high-level semantic-relational concepts that can be inferred from the low-level factor graph.
Given a set of mapped \textit{Planes} our algorithm is capable of inferring \textit{Room} entities relating to the \textit{Planes}.  
Additionally, to demonstrate the versatility of our method, our algorithm can infer an additional semantic-relational concept, i.e. \textit{Wall}, and its relationship with its \textit{Planes}.
We validate our method in both simulated and real datasets demonstrating improved performance over two baseline approaches. Furthermore, we integrate our method into the \textit{S-Graphs+} algorithm providing improved pose and map accuracy compared to the baseline while further enhancing the scene representation. 







\end{abstract}
\section{Introduction}
\label{introduction}

\begin{figure}[!t]
    \centering
    \includegraphics[width=0.45\textwidth]{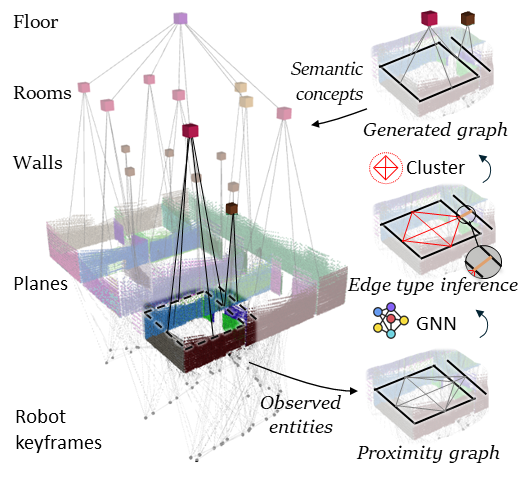}
    \caption{\textbf{System Overview.} We learn how to generate high-level semantic concepts, such as \textit{Rooms} and \textit{Walls} from low-level observed entities, such as \textit{Planes}. In this example, the \textit{Plane} information is retrieved from the low-level layer of \textit{S-Graphs+}~\cite{s_graphs+} and transformed in a \textit{proximity graph} in which a GNN classifies each edge as \textit{same Room} or \textit{same Wall}. The newly inferred edges are further clustered leveraging the existence of cycles, obtaining a new \textit{Room} or \textit{Wall} node for each cluster. By augmenting the \textit{S-Graphs+}~\cite{s_graphs+} with the new nodes and edges, we improve the quality of the map and the estimated camera trajectory.}
    \label{fig:front}
\end{figure} 
%


High-level semantic-relational entities enhance a robot's situational awareness~\cite{sa_survey} and enrich the built world model for improved scene understanding. 
It further provides advantageous information for successive tasks such as planning~\cite{paul} or robot navigation~\cite{huang23vlmaps}.

During recent years, 3D Scene Graphs \cite{3d_scene_graph, 3-d_scene_graph} have emerged as a promising framework to model the scene using semantic-relational concepts.  
Notably, \cite{sgf} takes a step further by generating the relations between observed objects in real time, although they do not include new entities. Hydra \cite{hydra} constructs and optimizes the 3D scene graph in real time leveraging loop closures.




Coupling them more tightly, \textit{S-Graphs+}~\cite{s_graphs+} generates a four-layered optimizable factor graph comprising a SLAM graph and a 3D scene graph as depicted in Fig.~\ref{fig:front}.  
The lowest layer of the graph contains the robot \textit{Keyframes} connected to the second layer, composed of directly observed raw geometric entities i.e.~\textit{Planes} (vertical planes named \textit{Walls} in~\cite{s_graphs+}).
The upper two layers represent a scene graph, containing semantic \textit{Room} entities relating with the underlying \textit{Planes} and semantic \textit{Floor} entities connecting with the respective \textit{Rooms}. 
However, \textit{Hydra}~\cite{hydra} and \textit{S-Graphs+}~\cite{s_graphs+} extracts these semantic \textit{Room} and \textit{Floor} entities using ad-hoc hand-tuned algorithms, thus not generalizable to complex and irregularly shaped indoor environments. 



To address these limitations, we present a framework to enhance the relational and generalization capabilities of 3D scene graphs by learning to generate semantic entities based on GNNs~\cite{wu2020comprehensive, velivckovic2017graph}. With it, we not only improve the state of the art of \textit{Room} generation but also generate new entities such as \textit{Walls} (two parallel vertical planes), all by a common architecture.
As shown in Fig.~\ref{fig:front}, the observed entities are transformed into a proximity graph used for message passing. The GNN classifies the type of each node into concepts of \textit{same Room}, \textit{same Wall} or none, which are further processed and clustered leveraging the existence of cycles. For each cluster, a new node of \textit{Room} or \textit{Wall} is generated along with the factors that tightly couple them with the underlying \textit{Planes}. Nodes and factors are incorporated back to the \textit{S-Graphs+}~\cite{s_graphs+} to validate its usefulness in semantic-relational SLAM.

As a ground truth graph for training the underlying GNN model, we first generate a synthetic dataset comprising of low-level entities i.e. \textit{Planes} and the higher-level semantic entities i.e. \textit{Rooms} and \textit{Walls} along with their relationships.
\
With no further fine-tuning of the trained GNN model, the results over several simulated and real structured indoor environments demonstrate that our method improves the baselines in detection time, expressiveness, and the number of entities detected. 


To summarize, the primary contributions of our paper are:

\begin{itemize}
	\item A GNN-based framework to generate high-level semantic entities (i.e. \textit{Rooms} and \textit{Walls}) and their relationships with the low-level entities (i.e. \textit{Planes}) in a precise, fast, and versatile manner.
	\item Integration of the algorithm within the four-layered optimizable \textit{S-Graphs+} framework \cite{s_graphs+} along with validation in simulated and real datasets with relevant ablations.
\end{itemize}

\section{Related work}
\label{related_work}
\subsection{Semantic Scene Graphs for SLAM}

Scene graphs serve as graph models that encapsulate the environment as structured representations. This graph comprises entities, their associated attributes, and the interrelationships among them. In the context of 3D scene graphs,\cite{3d_scene_graph} has pioneered the development of an offline, semi-autonomous framework. This framework relies upon object detections derived from RGB images, creating a multi-layered hierarchical representation of the environment and its constituent elements such as cameras, objects, rooms, and buildings. 3D DSG ~\cite{dynamic_scene_graph} extends this model to account for dynamic entities as humans in the scene. Furthermore,~\cite{sgf} segment instances, their semantic attributes, and the concurrent inference of relationships, in real time. \cite{gu2023conceptgraphs}, \cite{koch2024open3dsg} generate open-vocabulary 3D scene graphs by using open-vocabulary object detections and querying their relationships to suitable Large Language Models (LLMs). 
On the one hand, while the above 3D scene graph frameworks run SLAM/pose estimation backend, they do not utilize the generated scene graph to enhance the SLAM process and on the other hand, they can only infer nearby object relationships and are unable to estimate higher-level entities like \textit{Rooms} and its interconnections with the objects inside.  



Hydra\cite{hydra} focuses on real-time 3D scene graph generation performing a real time room segmentation and interconnecting the objects lying within the rooms while utilizing this information to enhance the loop closure search to finally optimize the entire scene graph. The extension of Hydra in~\cite{hughes2023foundations} introduces \textit{H-Tree} \cite{talak2021neural} to characterize the room detection to specific building areas, like kitchens, living rooms, etc. Both the above approaches do not completely integrate the scene graph elements within the SLAM state for simultaneous optimization and utilize an ad-hoc free-space voxel \cite{oleynikova2018sparse} based clustering for room identification, leading to misclassification of room entities in the presence of complex environmental setups. 


\textit{S-Graphs}\cite{s_graphs, s_graphs+}, creates a four-layered hierarchical optimizable graph performing real time room and floor segmentation while concurrently representing the environment as a 3D scene graph. However, the detection of the \textit{Room} entities is also performed using an ad-hoc free-space clustering approach based on \cite{oleynikova2018sparse} limiting its generalizability in different environments. 
\cite{greve2023collaborative} present a 3D scene graph construction for outdoor environment. Although using panoptic detector for detecting object instances, they utilize similar heuristics to extract high-level information about roads and intersections corresponding to rooms and corridors in the indoor 3D scene graphs. 

Analyzing the state-of-the-art regarding 3D scene graphs necessitates the requirement of a generic framework for the identification of higher-level semantics like \textit{Rooms}. Thus, to augment the reasoning capability of 3D scene graphs through efficient extraction of high-level semantic concepts and relating them to their low-level counterparts, we present a GNN-based framework integrated within the \textit{S-Graphs+} \cite{s_graphs+}, to infer high-level semantic concepts (\textit{Rooms} and \textit{Walls}) for a given set of low-level entities (\textit{Planes}).


\begin{figure*}[!t]
    \centering
    \includegraphics[width=1\textwidth]{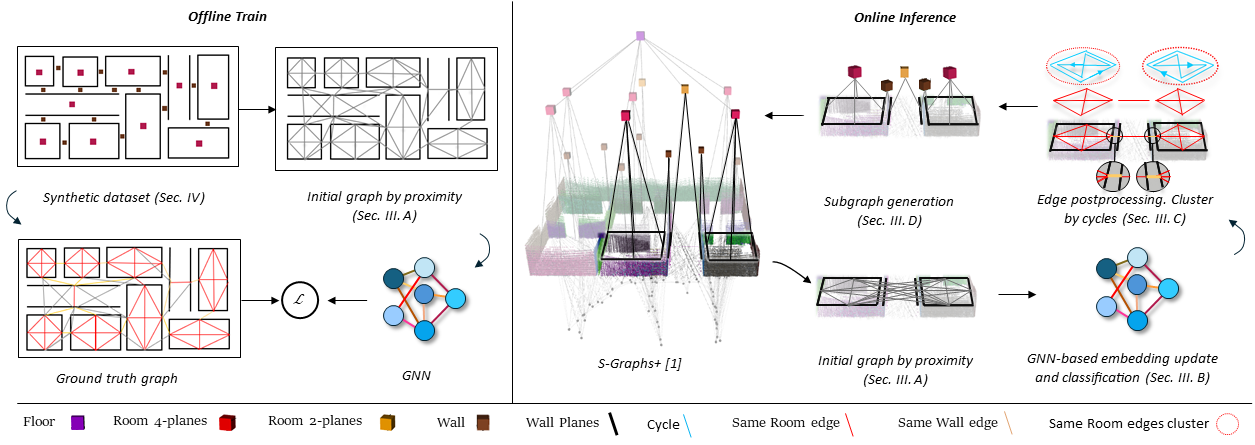}
    \caption{ \textbf{System Architecture.} This illustrates the entire process from geometric entities reception to the inclusion of new higher-level entities to \textit{S-Graph}. First, the GNN is trained off-line to update the embedding of nodes and edges and classify the edges. During the SLAM process, the raw low-level nodes are retrieved from \textit{S-Graphs} and connected with a proximity graph. The GNN infers edge classification to \textit{``same Room"}, \textit{``same Wall"}, or nothing. \textit{``same Room"} edges are clustered leveraging cycles. A subgraph is generated for clusters or standalone edges and finally included in \textit{S-Graph}.}
    \label{fig:general_scheme}
\end{figure*}

\subsection{Room and Wall Detection}

The first step in the generation of higher-level concepts resides in comprehending the interrelations among fundamental geometric entities. The identification of structural configurations corresponding to \textit{Planes} which collectively form \textit{Rooms} and \textit{Walls}, is crucial. Various methods have been explored to address this challenge, encompassing the utilization of pre-existing 2D LiDAR maps\cite{roomsegmentationreview, maoris, mapsegnet}, the utilization of 2D occupancy maps within complex indoor environments\cite{rose2}, and pre-established 3D maps\cite{3D_semantic_parsing, automatic_room_segmentation, 3d_room_recontruction}. It should be noted, however, that these approaches exhibit inherent performance constraints and lack real-time operational capabilities.\cite{hydra} introduce a real-time \textit{Room} segmentation approach using free-space clusters \cite{oleynikova2018sparse} designed to classify different places into \textit{Rooms}. \cite{s_graphs+} leverages the \textit{Planes} surrounding a given free-space cluster to instantaneously define \textit{Rooms} in real-time. To the best of our knowledge, no analogous methodologies exist based on GNNs to identify both \textit{Room} and \textit{Wall} entities for a given set of \textit{Planes}.

\section{Methodology}
\label{methodology}


The pipeline of our method is illustrated in  Fig.~\ref{fig:general_scheme}.
It can be mainly divided into two parts, offline training and online inference. Offline training utilizes a synthetically generated dataset (see Sec.~\ref{data_generation}) which contains an initial connected graph (see Sec.~\ref{sec:methodology_initial_graph}) based on proximity and the ground truth graph labels for sets of \textit{Planes} forming either \textit{``same Room"} or \textit{``same Wall"} relations. Both the initial graph and the ground truth labels are utilized to train a GNN model. 

In the part of online inference, first, the mapped \textit{Plane} features from the low-level layer of \textit{S-Graphs+}~\cite{s_graphs+} are received. 
These \textit{Plane} features are preprocessed to build a proximity graph and define the initial embedding of nodes and edges (see Sec.~\ref{sec:methodology_initial_graph}) which serve as an input to the trained GNN model.
The trained GNN model updates these initial edge embeddings between the nodes, classifying them appropriately into either \textit{``same Room"} or \textit{``same Wall"} relations, further explained in Sec.~\ref{sec:methodology_gnn}. 
Furthermore, these inferred edges are appropriately clustered to generate new nodes of either \textit{Wall} or \textit{Room} entities appropriately connected with the underlying \textit{Planes} (Sec.~\ref{sec:methodology_subgraph_generation}).  
Finally, these new nodes and their relationships are integrated into the high-level \textit{Rooms} and \textit{Walls} layer of \textit{S-Graphs+}~\cite{s_graphs+}.

\subsection{Initial Graph by Proximity}
\label{sec:methodology_initial_graph}

The \textit{Plane} features before being input to the GNN are preprocessed to generate an initial graph based on the proximity of these features. 
This module is utilized in both offline training and online inference (see Fig.~\ref{fig:general_scheme}). In case of offline training, the initial \textit{Plane} features in the synthetic dataset are defined as $\boldsymbol{\pi'}_{i} = [\boldsymbol{n}_{i}, {w}_{i}, c_i]$, where $\boldsymbol{n}_{i}$ is a normal orientation of the plane defined in closest point form as in \cite{s_graphs+}, ${w}_{i}$ and $c_i$ are the width and the centroid of the plane.  

In online inference, \textit{Plane} features are received from \textit{S-Graphs+} \cite{s_graphs+} in the closest point form as $\boldsymbol{\pi}_{i} = [\boldsymbol{n}_{i}, d_i]$, $\boldsymbol{n}_{i}$ being the normal orientation and $d_i$ being the perpendicular distance to the origin. 
Furthermore, each \textit{Plane} feature
also includes the set of 3D points $\boldsymbol{p}_{i}\in\mathbb{R}^3$ from the observed point cloud. 
To obtain the width and centroid of these \textit{Planes} we first flatten and assimilate all the 3D points to a 2D line segment. 
Subsequently, due to noise in the plane mapping step of \textit{S-Graphs+}~\cite{s_graphs+}, there could be the presence of duplicate \textit{Planes} and same \textit{Planes} could be shared between different \textit{Rooms}. To overcome this, we first filter out duplicate planes and then split the \textit{Planes} based on their 2D line intersections with the neighboring \textit{Planes}. 


At this point, we have \textit{Planes} represented as $\boldsymbol{\pi'}_{i}$ in both offline training and online inference.  
Finally, as described in Fig.~\ref{fig:concepts_and_embedings}b, the initial \textit{Plane} embedding $\boldsymbol{v}^0_{i}$ for the GNN is defined as $[\boldsymbol{n}_{i}, w_{i}]$. 
Given the centroid information for each \textit{Plane} feature new directed edges are generated based on their proximity to the other \textit{Planes}. The embedding for these edges can be defined as  $e^0_{ij} = [\delta(c_{j}, c_{i}), cd_{ij}]$,  being $\delta(c_{j}, c_{i})$ the relative position of the centroids of ${i}$ and $j$ \textit{Planes} and $cd_{ij}$ being the closest distance between their segment extremes.


\subsection{GNN-based Embedding Update and Classification}
\label{sec:methodology_gnn}

\begin{figure}[!t]
\centering
\small
\setlength{\tabcolsep}{2pt}
\renewcommand{\arraystretch}{1.0}
\newcommand{\sz}{0.47}
\begin{tabular}{c|c}
  \includegraphics[width=0.4\columnwidth]{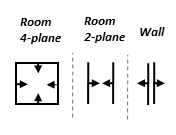}  & 
  \includegraphics[width=0.4\columnwidth]{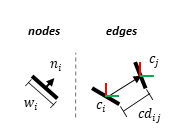}  \\
   (a) Learned Patterns & (b) Initial Embeddings\\[6pt]
  \hline
  \multicolumn{2}{c}{
  \includegraphics[width=0.85\columnwidth]{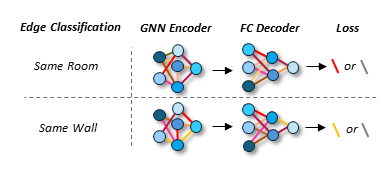}} \\
  \multicolumn{2}{c}{(c) Neural Architecture}\\
\end{tabular}
\caption{a) \textbf{Learned Patterns.} Example of the distribution of \textit{Planes} that belong to the same high-level concept. b) \textbf{Initial Embeddings.} The definition of the nodes features of the \textit{Planes} is composed by the width (w) and the normal (n) from which it was observed. The edge features are defined by the relative position of the centroids (c) and the closest distance (cd). c) \textbf{Neural Architecture.} The classification of both \textit{``same Room"} and \textit{``same Wall"} relations is accomplished by two independent neural networks with similar architecture.
}
\label{fig:concepts_and_embedings}
\end{figure}

All the edges contained in the initial graph are classified into relations of either \textit{``same Room"}, \textit{``same Wall"}, or none by the GNN-based model trained in the offline training step. For each relation of \textit{``same Room"} and \textit{``same Wall"}, the classification is performed by two separate GNN models. 
As shown in Fig.~\ref{fig:concepts_and_embedings}c and inspired by~\cite{sgf}, both models have the same encoder-decoder architecture but have different hyperparameters.

The GNN-based encoder updates the node and edge embeddings separately but interleaved using the latest updates as below:
\begin{align}
v_{i}^{l+1} &= g_{v} \big(\big[v_{i}^l, \max_{j\in\mathcal{N}(i)}(\mathrm{GAT}(v_i^l,e_{ij}^l,v_j^l))\big]\big)
\label{eq:encoder_v}
\\
e_{ij}^{l+1} &= g_{e}\big([v_i^l,e_{ij}^l,v_j^l]\big)
\label{eq:encoder_e}
\end{align}
\noindent
where $g_{v}(\cdot)$ and $g_{e}(\cdot)$ are linear layers~\cite{murtagh1991multilayer}, $\mathcal{N}(i)$ are the neighbors of $i_{th}$ node and $\mathrm{GAT}(\cdot)$ is a Graph Attention Network~\cite{wu2020comprehensive,velivckovic2017graph}. Encoder hyperparameters are maintained across the classification of both relations. 
Eq.~\eqref{eq:encoder_v} and Eq.~\eqref{eq:encoder_e} utilize two hidden layers. 

The latest embeddings from the encoder are passed through a multi-layer perceptron decoder as follows:
\begin{equation}
c_{ij} = g_{d}([v_i^L,e_{ij}^L,v_j^L])
\label{eq:decoder}
\end{equation}

\noindent
where $g_{d}(\cdot)$ are three linear layers and $L$ is the last layer of the encoder and $c_{ij}$ is the final binary classification of a specific edge.


\subsection{Edge Postprocessing}
\label{sec:methodology_edge_postprocessing}

We further post-process the binary classification of the edges using appropriate clustering to account for misclassifications from the GNN and  select \textit{Planes} that belong to the same \textit{Wall} and \textit{Room} entities.  
Since a \textit{Wall} node is only related to two \textit{Planes} (see Fig.~\ref{fig:concepts_and_embedings}a), only one \textit{``same Wall"} edge is required, avoiding the need for further clustering.

For \textit{``same Room"} relation, the existence of cycles is leveraged.
We assume all \textit{Plane} nodes forming a \textit{``same Room"} relationship are connected through at least one cycle and a \textit{Plane} only relates to one \textit{Room}.
For each cycle, a \textit{Plane} set is obtained. We prioritize those sets by the following criteria: (1) sets with largest number of \textit{Planes} and (2) highest repetitions of the same \textit{Plane} set.
These criteria overcome the issue of the existence of false positives that may lead to the classification of the same \textit{Planes} for two different \textit{Rooms}. In our current implementation, we extract cycles of either two or four \textit{Planes} relating to a \textit{``same Room"} relation (i.e. 2-Plane or 4-Plane Rooms). 

\subsection{Subgraph Generation}
\label{sec:methodology_subgraph_generation}

Finally, after performing the edge post-processing step and filtering out unwanted/misclassified edges, we are left with edge sets relating \textit{Planes} as either \textit{``same Room"} or \textit{``same Wall"}. Based on the inferred relation type we can generate either \textit{Room} or \textit{Wall} nodes. The center of a \textit{Room} node can be computed as follows:
\begin{equation}
\boldsymbol{\rho}_i = \frac{\sum_{j=1}^{\mathcal{N}_i} c_j}{\mathcal{N}_i}
\label{eq:centers}
\end{equation}
\noindent where centroid $c_j$ is the centroid of a given \textit{Plane} and $\mathcal{N}_i$ is the set of all planes connected to the \textit{Room} node. \textit{Wall} node center can be computed in the same manner following  
Eq.~\eqref{eq:centers}. 

These newly generated nodes along with centers and their connected \textit{Planes} are incorporated into the \textit{Rooms} and \textit{Walls} layer of the optimizable factor graph of \textit{S-Graphs+} \cite{s_graphs+}. 
The cost function for \textit{Room} with center $\boldsymbol{\rho_i}$ and its four \textit{Planes} $\boldsymbol{\pi}_n$ can be given as: 
\begin{multline} \label{eq:finite_room_node}
    c_{\boldsymbol{\rho}} ({\boldsymbol{\rho}}_i, \big[ {\boldsymbol{\pi}_1}, {\boldsymbol{\pi}_2}, {\boldsymbol{\pi}_3}, {\boldsymbol{\pi}_4}\big]) = \\ \| {\hat{\boldsymbol{\rho}_i}} - {{f({\hat{\boldsymbol{\pi}}_1}, {\hat{\boldsymbol{\pi}}_2}, {\hat{\boldsymbol{\pi}}_3}, {\hat{\boldsymbol{\pi}}_4})}} \| ^2_{\mathbf{\Lambda}}
\end{multline}
\noindent $f(\cdot)$ is the function that maps the room center using the four \textit{Planes}. 

The cost function for a \textit{Room} $\boldsymbol{\kappa}_i$ with its two \textit{Planes} $\boldsymbol{\pi}_n$ can be given as:
\begin{equation} \label{eq:infinite_room_node}
    c_{\boldsymbol{\kappa}_i}({\boldsymbol{\kappa}_i},\big[{\boldsymbol{\pi}_1}, {\boldsymbol{\pi}_2}, {\textbf{p}_i}\big]) \\ = \| {\hat{\boldsymbol{\kappa}}_i} - f({\hat{\boldsymbol{\pi}}_1}, {\hat{\boldsymbol{\pi}}_2}, {\textbf{p}_i}) \| ^2_{\mathbf{\Lambda}}
\end{equation}
\noindent where ${\textbf{p}_i} \in \mathbb{R}^3$ is the average of the centroid of the two planes and  $f(\cdot)$ is the function that maps the room center using the two \textit{Planes} and points ${\textbf{p}_i}$. \textit{Wall} node is incorporated into the factor graph using cost $c_{\boldsymbol{w}_i}$ which follows the same Eq.~\ref{eq:infinite_room_node}. In both Eq.~\ref{eq:finite_room_node} and Eq.~\ref{eq:infinite_room_node}, $\hat{\scriptstyle(\cdot)}$ refers to the estimated values of the variables in the optimizable graph. 




\section{Training with synthetic dataset}
\label{data_generation}

We generate synthetic dataset that provides ground truth labels of relationships between target entities, avoiding the need for extracting and labelling real world datasets. 
We developed the synthetic dataset focusing on replicating common \textit{Plane} structure of usual indoor environments. 
As explained in Sec.~\ref{sec:methodology_initial_graph}, the \textit{Planes} $\boldsymbol{\pi'}_{i}$ is defined in closest point form with its width and centroid. The dataset also contains the center of the \textit{Rooms} $\boldsymbol{\rho}_i$ and \textit{Walls} $\boldsymbol{w}_i$ appropriately relating to the underlying \textit{Planes}. 
Furthermore this data is augmented with several layers of randomization in size, position, and orientation when creating \textit{Rooms} and \textit{Planes}. 
Ground truth edges between \textit{Planes} are automatically tagged as \textit{same Room} or \textit{same Wall} concepts 
and included along with negative tagged edges with the 15 closest neighbor \textit{Plane} nodes for a given \textit{Plane} node. During the training process, 800 different layouts are used for backpropagation during each one of the 35 epochs. Xavier uniform initialization \cite{xavier} is used for the learnable parameters. 

One of the advantages of our approach is that, after the initial training with the synthetic dataset, the trained GNN model is used on real data without further retraining, with no additional tuning of parameters and the same applied normalization. This gives the flexibility of training our model for different kinds of complex environment without the need for tedious data collection and labelling process.  

%
%


\section{Experimental Results}
\label{results}

\subsection{Methodology}
Our work is validated in both simulated and real datasets presented in \cite{s_graphs+} collected using a Velodyne VLP-16 3D LiDAR over different indoor environments comprising office spaces and constructions sites detailed in~Tab.~\ref{tab:scenes}.  
We compare the graph expressiveness through precision and recall of our algorithm with the ad-hoc \textit{Room} detection algorithms presented in Hydra~\cite{hydra} and \textit{S-Graphs+} \cite{s_graphs+}, which we call \textit{Hydra RS}  and \textit{S-Graphs+ RS} respectively. In addition, our algorithm is tested for its precision and recall in both Conservative (C) or Greedy (G) scenarios for \textit{Room} generation. In the \textit{Ours G} scenario lower threshold value is applied in the \textit{same Room} edge classification type when compared to the \textit{Ours C} scenario. We also compare the First Detection Time of our approach to the room segmentation of the baselines. 


Furthermore, we integrate our algorithm within the \textit{S-Graphs+} \cite{s_graphs+} framework replacing its ad-hoc room segmentation algorithm while naming it \textit{Ours~(Int.)}. 
We compare the pose and map accuracy of \textit{Ours~(Int.)} with the baseline \textit{S-Graphs+} \cite{s_graphs+} algorithm. 
Additionally, we ablate \textit{Ours~(Int.)} method incorporating only \textit{Rooms} without \textit{walls} calling it \textit{Ours~(Int.) (rooms only)}. Given the lower variance in precision/recall for \textit{Ours C} approach, we choose it for in the \textit{Ours (Int.)} approach for further validations of ATE and MMA.  
The example scenarios are preceded with \textit{S} and \textit{R} to differentiate simulation and real datasets respectively. 

In all the experiments, no fine-tuning of the specified network hyper-parameters is applied, as the empirically chosen ones during the training (Sec.~\ref{data_generation}) suffice for all cases.

\boldparagraph{Simulated Data.} We performed a total of five experiments on simulated datasets denoted as \textit{SC1F1}, \textit{SC1F2}, \textit{SE1}, \textit{SE2}, and \textit{SE3}. \textit{SC1F1} and \textit{SC1F2} are generated from the 3D meshes of two floors of actual architectural plans, while \textit{SE1}, \textit{SE2} and \textit{SE3} simulate typical indoor environments with varying \textit{Room} configurations. In these experiments we compute the graph expressiveness of our approach and baselines in terms of precision/recall for \textit{Room} and \textit{Wall} detection given the availability of ground truth \textit{Rooms} and \textit{Walls}. 

Additionally, In order to assess the pose and map accuracy of \textit{Ours~(Int.)} with the \textit{S-Graphs+} \cite{s_graphs+} baseline we report Average Trajectory Error (ATE) and Map Matching Accuracy (MMA). The ATE is calculated against the ground truth provided by the simulator and the MMA is computed utilizing the ground truth map available for each scenario.

\boldparagraph{Real Dataset.} We conducted four real experiments in two different construction sites. \textit{RC1F1} and \textit{RC1F2} are conducted on two floors of a small construction site. \textit{RC2F2} and \textit{RC3F2} are conducted in two other construction sites with larger areas. 
First, we validate the graph expressiveness of \textit{Rooms} and \textit{Walls} detection of our approach and baselines with the ground truth \textit{Rooms} and \textit{Walls}. 
Second, to validate the accuracy of each method in all real-world experiments, we report the MMA of the estimated 3D maps in comparison to the ground truth 3D map generated from the architectural plans. 

\begin{table}[tb]
\caption{\textbf{Scenes Description.} Enumeration of all simulated and real scenes included in the validation. \textit{Room} shapes can be squared, L-shaped, elongated, or corridors.
}
\centering
\footnotesize
\setlength{\tabcolsep}{1.9pt}
\renewcommand{\arraystretch}{1.2}
\begin{tabular}{c | c | c}
\toprule
\textbf{Scene} & World & Description \\
\midrule
SC1F1 & Simulated & 1 L-shaped, 1 squared rooms and 2 corridors. \\
SC1F2 & Simulated & 5 squared, 2 elongated rooms and 1 corridor. \\
SE1 & Simulated & 6 squared rooms and 3 corridors. \\
SE2 & Simulated & 5 squared and 2 elongated rooms. \\
SE3 & Simulated & 22 squared rooms and 4 corridors. \\
RC1F1 & Real & 1 L-shaped, 1 squared rooms and 2 corridors. \\
RC1F2 & Real & 5 squared, 2 elongated rooms and 1 corridor. \\
RC2F2 & Real & 7 squared, 2 L-shaped and 3 elongated rooms. \\
RC3F2 & Real & 9 squared, 3 L-shaped rooms and 2 corridors. \\
\bottomrule 
\end{tabular}
\label{tab:scenes}
\end{table}

\subsection{Results and Discussion}

\begin{figure}[h]
\centering
\small
\setlength{\tabcolsep}{2pt}
\renewcommand{\arraystretch}{1.0}
\newcommand{\sz}{0.47}
\begin{tabular}{ccc}
  \includegraphics[width=0.42\columnwidth]{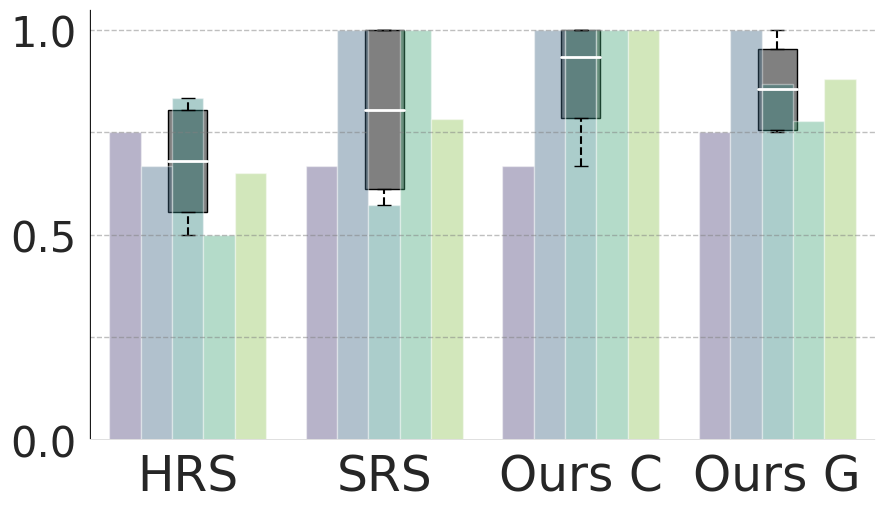} &
  \includegraphics[width=0.42\columnwidth]{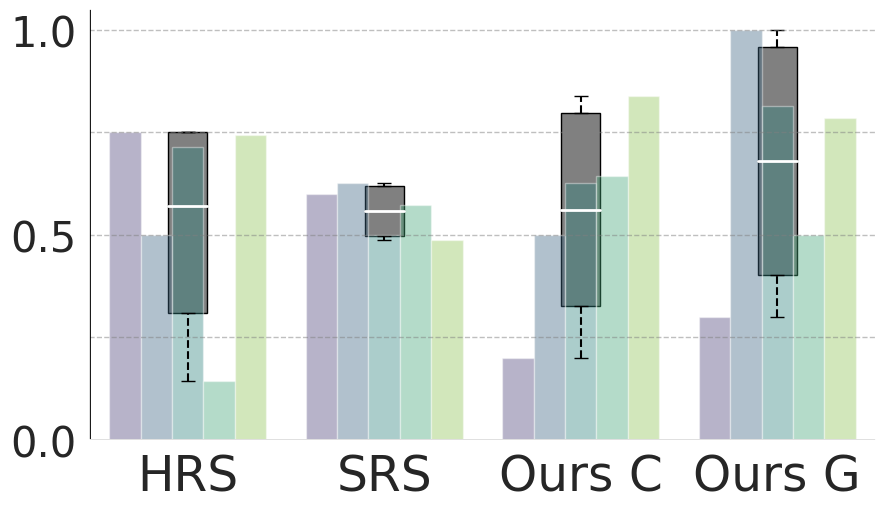} &
  \includegraphics[width=0.095\columnwidth]{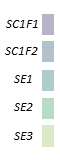} \\
  (a) Simulation, precision & (b) Simulation, recall\\[6pt]
  \includegraphics[width=0.42\columnwidth]{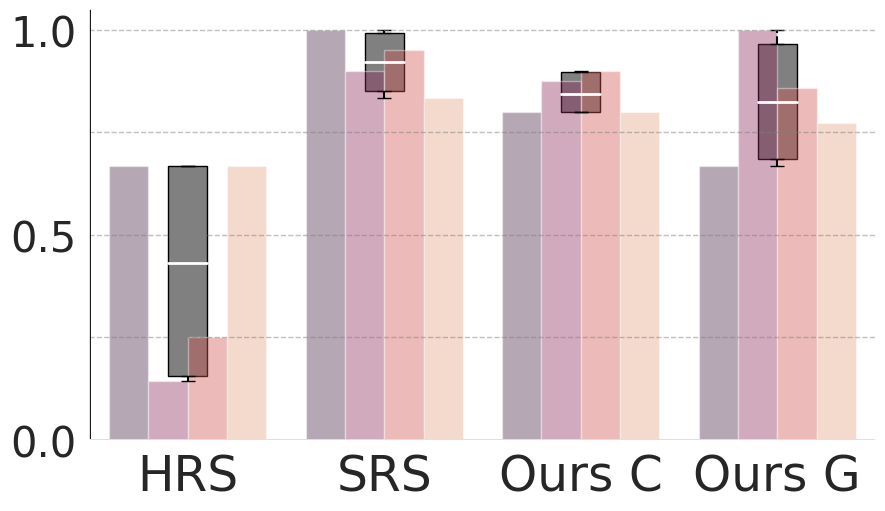} &
  \includegraphics[width=0.42\columnwidth]{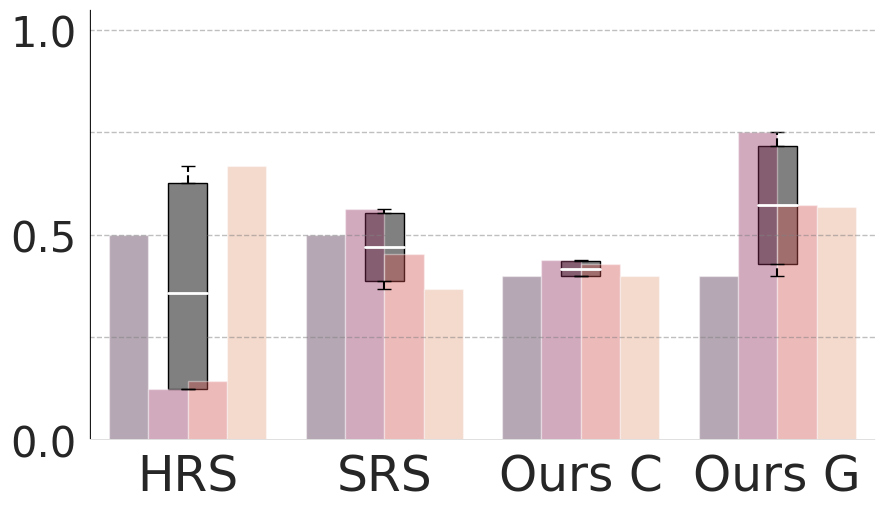} &
  \raisebox{0.2\height}{\includegraphics[width=0.095\columnwidth]{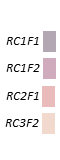}} \\
  (c) Real, precision & (d) Real, recall\\
\end{tabular}
\caption{\textbf{Graph Expressiveness.} measured by precision and recall in \textit{Room} detection for the in \textit{Hydra RS}~\cite{hydra} (HRS) and in \textit{S-Graphs+ RS} ~\cite{s_graphs+} (SRS), as baselines. Those are compared with \textit{Our} Conservative \textit{(C)} and Greedy \textit{(G)} approach in five simulated and four real scenes. For each approach, experiments are aggregated and the metrics are presented by mean, std, maximum, and minimum. 
}
\label{fig:precision_recall}
\end{figure}

\boldparagraph{Graph Expressiveness.}
Fig.~\ref{fig:precision_recall} showcases the precision/recall performance on the detection of \textit{Rooms} for simulated and real scenarios. Room segmentation algorithms used in \textit{Hydra}~\cite{hydra} and \textit{S-Graphs+}~\cite{s_graphs+} are compared with our ablated module \textit{(rooms only)}. Our ablation also assesses the relaxation of the GNN threshold to classify \textit{same Room} edges i.e Greedy (G) and Conservative (C) approach. 

For simulated/real scenarios, \textit{Ours C} approach improves the average precision over \textit{Hydra}~\cite{hydra} by $37\%/9.5\%$ and the recall by $0\%/16\%$ in simulated and real scenarios respectively.
With respect to \textit{S-Graphs+}~\cite{s_graphs+}, \textit{Ours C} provides the precision in simulated/real scenarios as $16\%/0\%$. In terms of recall, \textit{Ours C} maintains the same average recall in simulated scenarios while is in real scenarios the average recall deteriorates by $11\%$. 
Finally, although \textit{Ours G} with respect to \textit{Ours C} provides average improvements in recall by $21\%/37\%$, \textit{Ours G} degrades in average precision by $-8\%/2\%$. 

Fig.~\ref{fig:rviz_examples} presents qualitative results of graph expressiveness for experiments SE3 and RC2F2. As can be seen from the figure, although in SE3 the performance of \textit{Hydra RS}~\cite{hydra} is maintained on the average including 4 false positives, in RC2F2, most of the rooms present points misplaced over the whole area given the complexity of the real environment and the noise in the LiDAR measurements. 
Note from the figure that while \textit{S-Graphs+ RS}~\cite{s_graphs+} is able to include \textit{2-Plane Rooms} (orange squares) in both simulated and real scenarios, \textit{Ours C} is able to provide segment higher quantity of \textit{4-Plane Rooms} (pink squares) in both simulated and real datasets with higher precision, additionally it is able to identity and segment the \textit{Wall} entities. 


On its side, \textit{Wall} segmentation can not be compared to these baselines as they do not segmented by them. Thus in case \textit{Wall} segmentation we present Tab.~\ref{tab:wall_detection} providing precision/recall results compared to the ground truth data. It is worth mentioning that precision is always maintained at $1.0$ across simulated and real scenarios. Recall is over $75\%$ in all scenarios but in a simulated one. On the contrary to \textit{Rooms}, \textit{Walls} present a similar structure, which simplifies the task of finding the patterns by the GNN thus providing better results.



\begin{table}[tb]
\setlength{\tabcolsep}{6.8pt}
\caption{\textbf{Graph Expressiveness.} measured by precision and recall for \textit{Wall} detection of our approach in different simulated and real scenes. 
}
\centering
\footnotesize
\renewcommand{\arraystretch}{1.2}
\begin{tabular}{l | l l l l l | c}
\toprule
& \multicolumn{3}{l}{\textbf{Dataset}} \\ 
\toprule
\textbf{Metric} & \textit{SC1F1} & \textit{SC1F2} & \textit{SE1} & \textit{SE2} & \textit{SE3} & Avg \\
\midrule
Precision & 1.00 & 1.00 & 1.00 & 1.00 & 1.00 &  1.00 \\
Recall & 0.80 & 1.00 & 0.87 & 0.43 & 0.89 &  0.80 \\ 
\toprule
 & \textit{C1F1r} & \textit{C1F2r} & \textit{C2F2r} & \textit{C3F2r} &  & Avg \\
 \midrule
 Precision & 1.00 & 1.00 & 1.00 & 1.00 &  & 1.0 \\
Recall & 0.80 & 0.75 & 0.82 & 0.84 &  &  0.80 \\ 
\bottomrule
\end{tabular}
\label{tab:wall_detection}
\end{table}  

\boldparagraph{First Detection Time.}
Tab.~\ref{tab:compute_time} provides a comprehensive overview of the time required by each module to accomplish detection of the first \textit{Room}. Every experiment is started inside the construction site, that is inside of a \textit{Room}.  \textit{Ours C} is compared with \textit{S-Graphs+ RS} \cite{s_graphs+}, demonstrating a drastic average improvement of $84.3\%$ in the simulated datasets and $62.7\%$ in the real datasets. This is due to the fact that the room segmentation of \textit{S-Graphs+} needs to observe more map points until a free-space cluster can be inferred and then associate the cluster with the mapped \textit{Planes}, while our method utilizing only the mapped \textit{Planes} as input succeeds in finding the rooms faster.

\begin{table}[tb]
\setlength{\tabcolsep}{1.7pt}
\caption{\textbf{First Detection Time (FDT)} [s] of \textit{Rooms} on simulated and real data. Our method is substantially faster than the baseline. The best results are boldfaced.}
\centering
\footnotesize
\renewcommand{\arraystretch}{1.2}
\begin{tabular}{l | l l l l l | c}
\toprule
& \multicolumn{3}{l}{\textbf{Dataset}} \\ 
\toprule
 & \multicolumn{5}{l}{\textbf{Computation Time} (mean) [ms]}  \\
\toprule
\textbf{Module} & \textit{SC1F1} & \textit{SC1F2} & \textit{SE1} & \textit{SE2} & \textit{SE3} & Avg \\
\midrule
\textit{S-Graphs+ RS}~\cite{s_graphs+} (baseline) & 76.0 & 19.0 & 19.0 & 115.0 &  160.0 &  77.8 \\
\textit{Ours C} & \textbf{2.7} & \textbf{2.8} & \textbf{10.2} & \textbf{8.2} & \textbf{37.1} &  \textbf{12.2}  \\ 
\toprule
 & \textit{RC1F1} & \textit{RC1F2} & \textit{RC2F2} & \textit{RC3F2} &  & Avg \\
 \midrule
 \textit{S-Graphs+ RS}~\cite{s_graphs+} (baseline) & 19.0 & 25.5 & 367.0 & 308.0 &  & 179.9 \\
\textit{Ours C} & \textbf{1.7} & \textbf{2.6} & \textbf{54.0} & \textbf{210.2} &  & \textbf{67.1} \\ 
\bottomrule
\end{tabular}
\label{tab:compute_time}
\end{table}  

\begin{figure*}[!thb]
  \centering
  \setlength{\tabcolsep}{2pt}
  \newcommand{\sz}{0.18}
  \newcommand{\sh}{3cm}
  \begin{tabular}{cc|c|cc}
    & \textit{Hydra RS}~\cite{hydra} & \textit{S-Graphs+ RS}~\cite{s_graphs+} & \multicolumn{2}{c}{\textit{Ours C}}\\
    \raisebox{2\height}{\rotatebox{90}{SE3}} &
    \includegraphics[height=\sh]{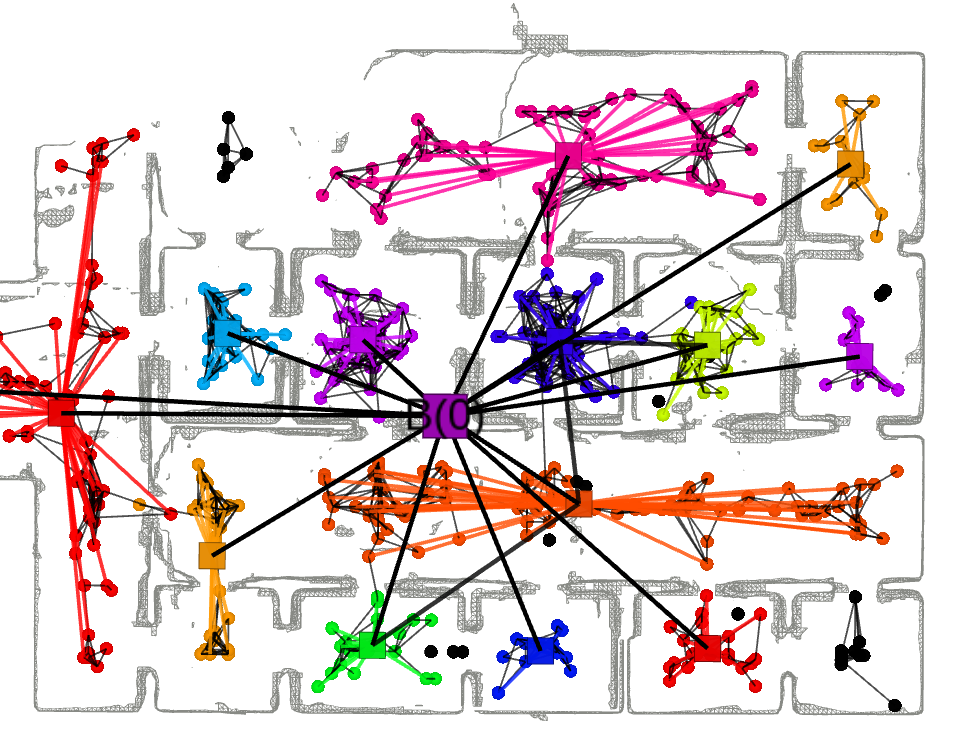} &
    \includegraphics[height=\sh]{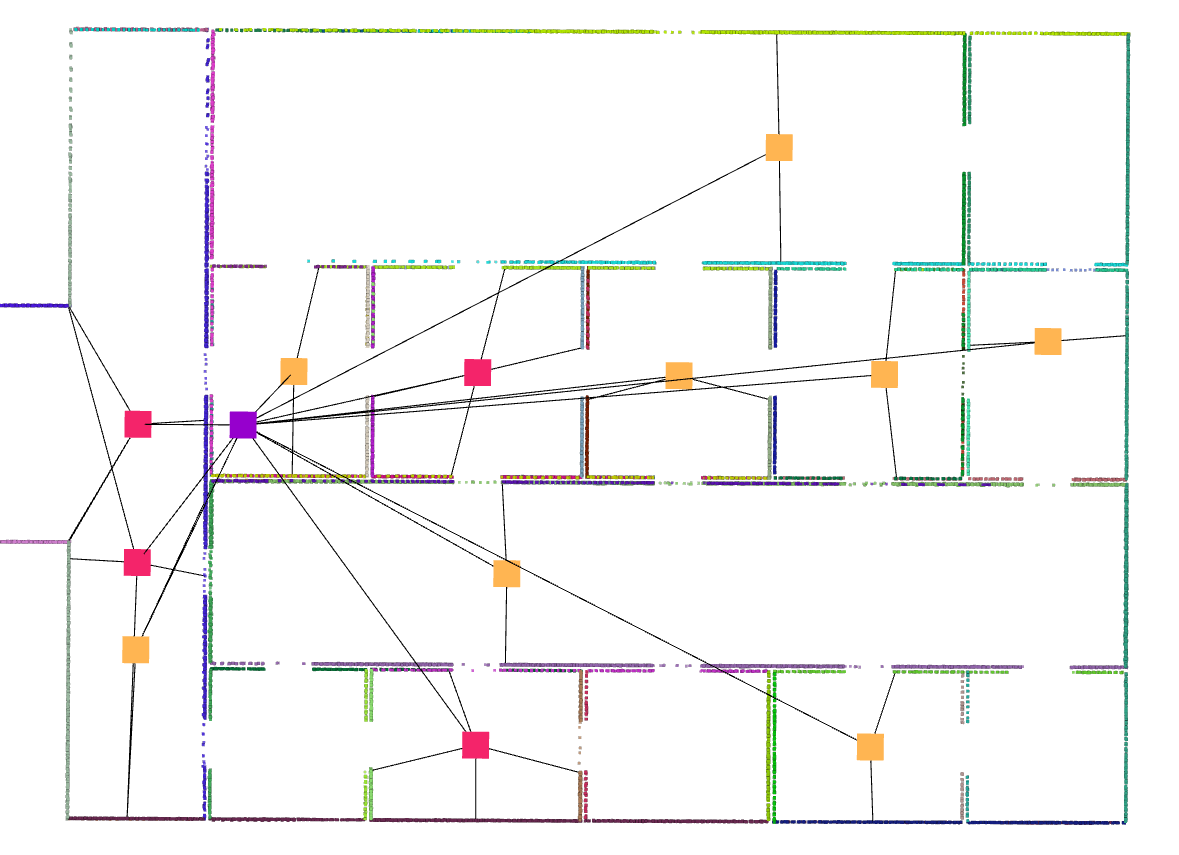} & 
    \includegraphics[height=\sh]{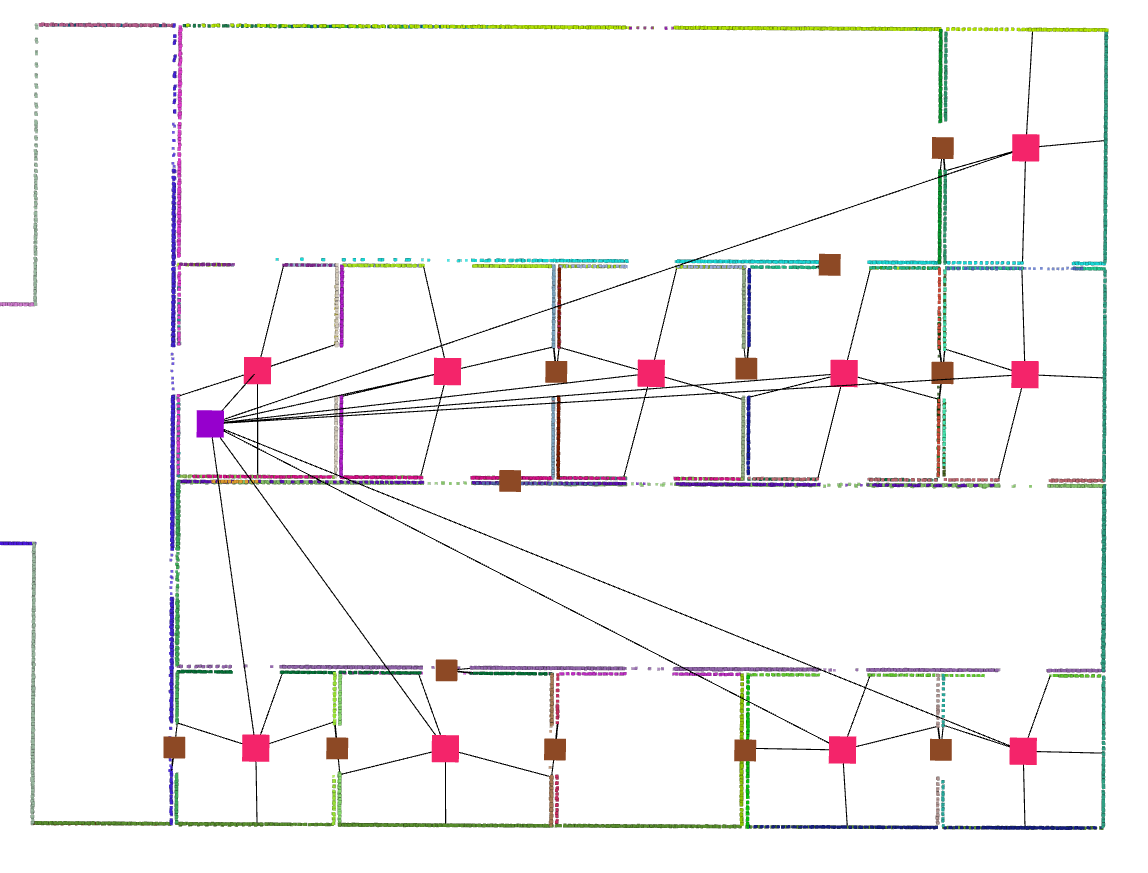} &
    \includegraphics[height=\sh]{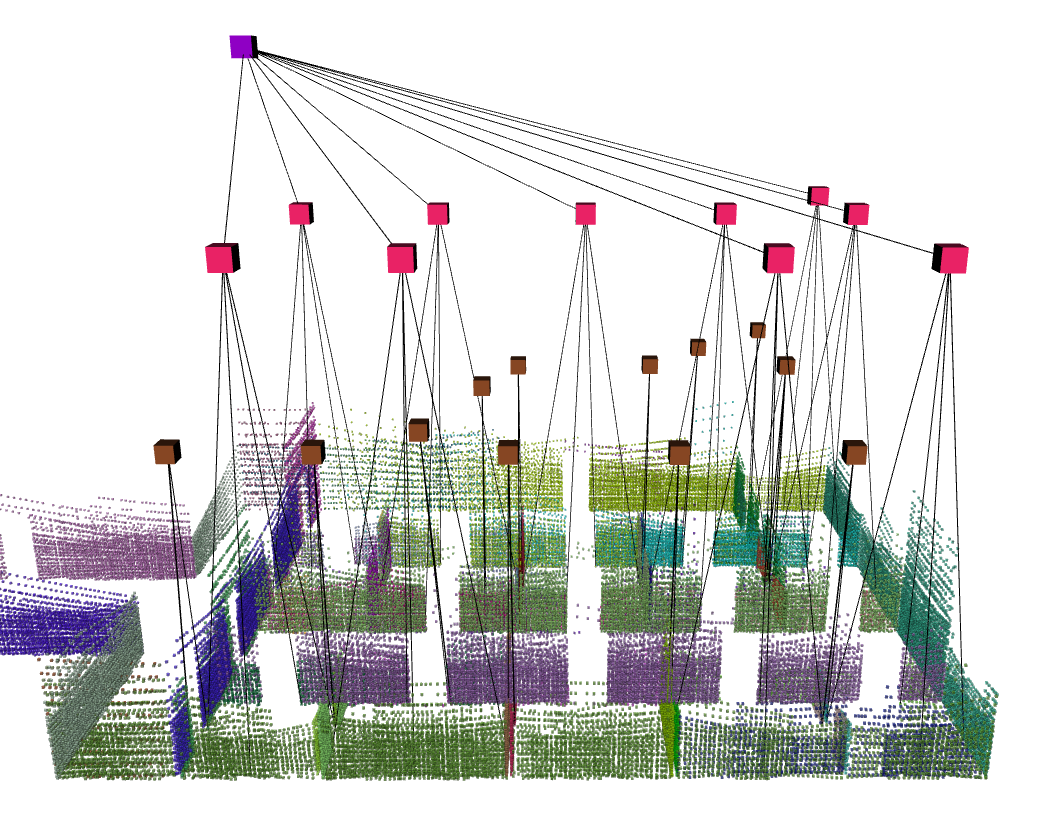} \\
    & (a) P: 9/13, R: 9/14 & (b) P: 7.5/9.5, R: 7.5/14 & (c) P: 10/10, R: 10/14 & (d) Walls: 12/12\\
    \raisebox{1\height}{\rotatebox{90}{RC2F2}} & 
    \includegraphics[height=\sh]{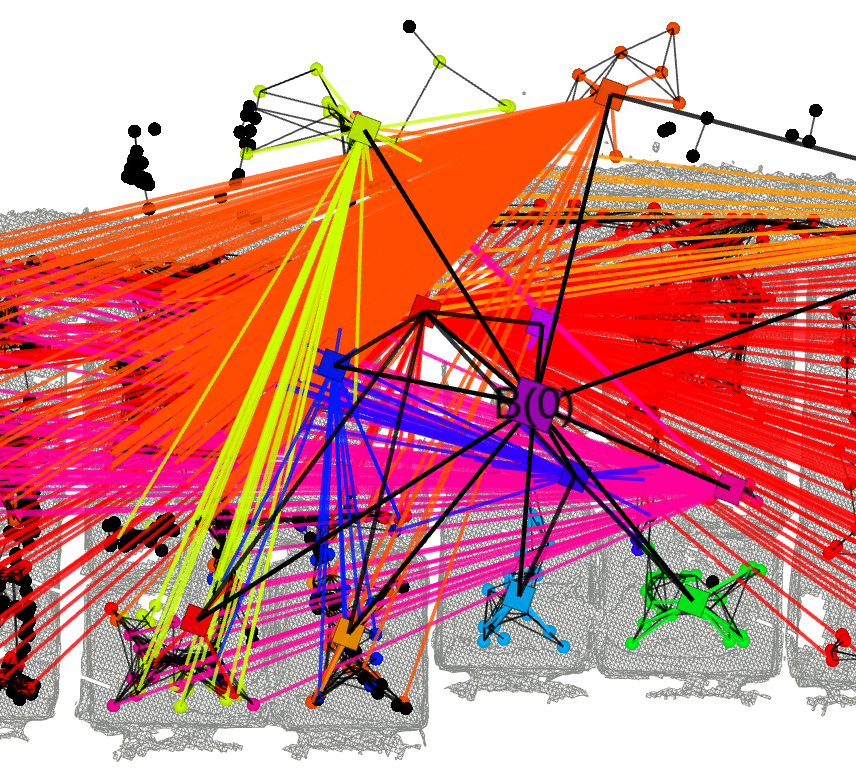} & 
    \includegraphics[height=\sh]{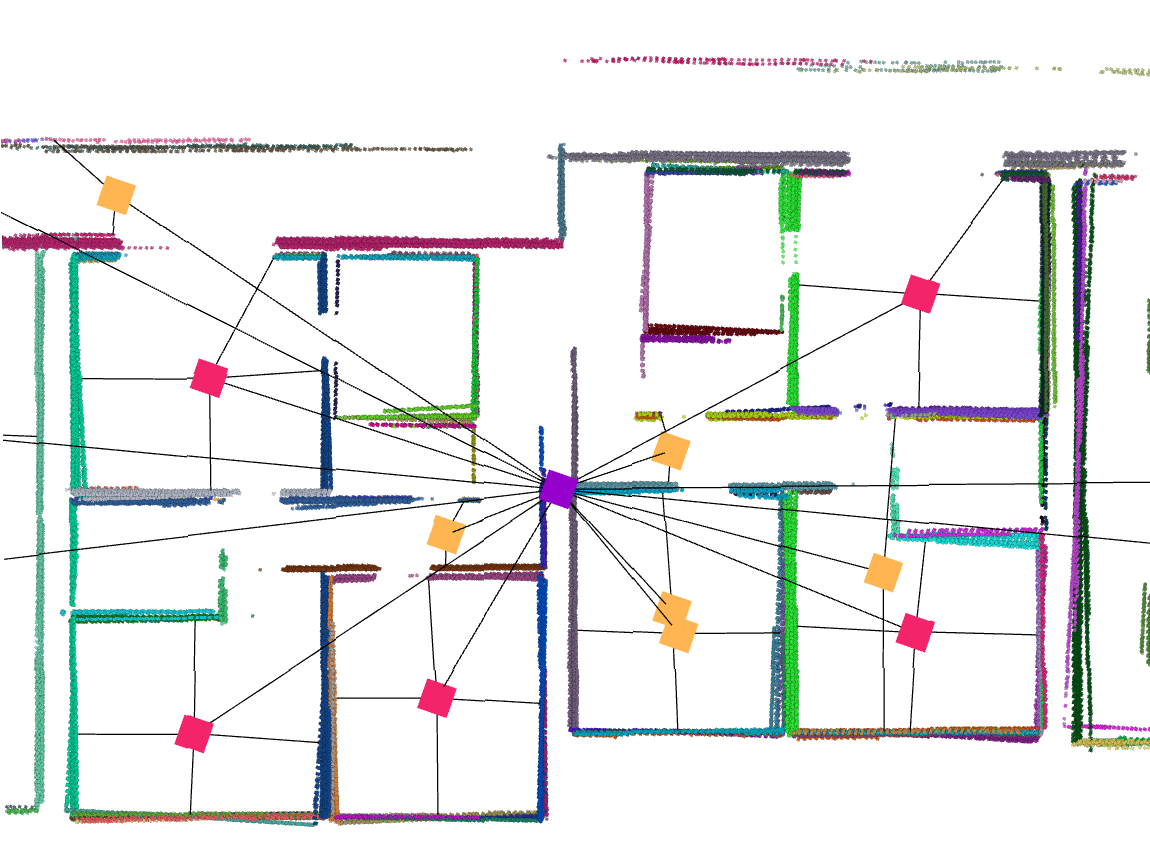} & 
    \includegraphics[height=\sh]{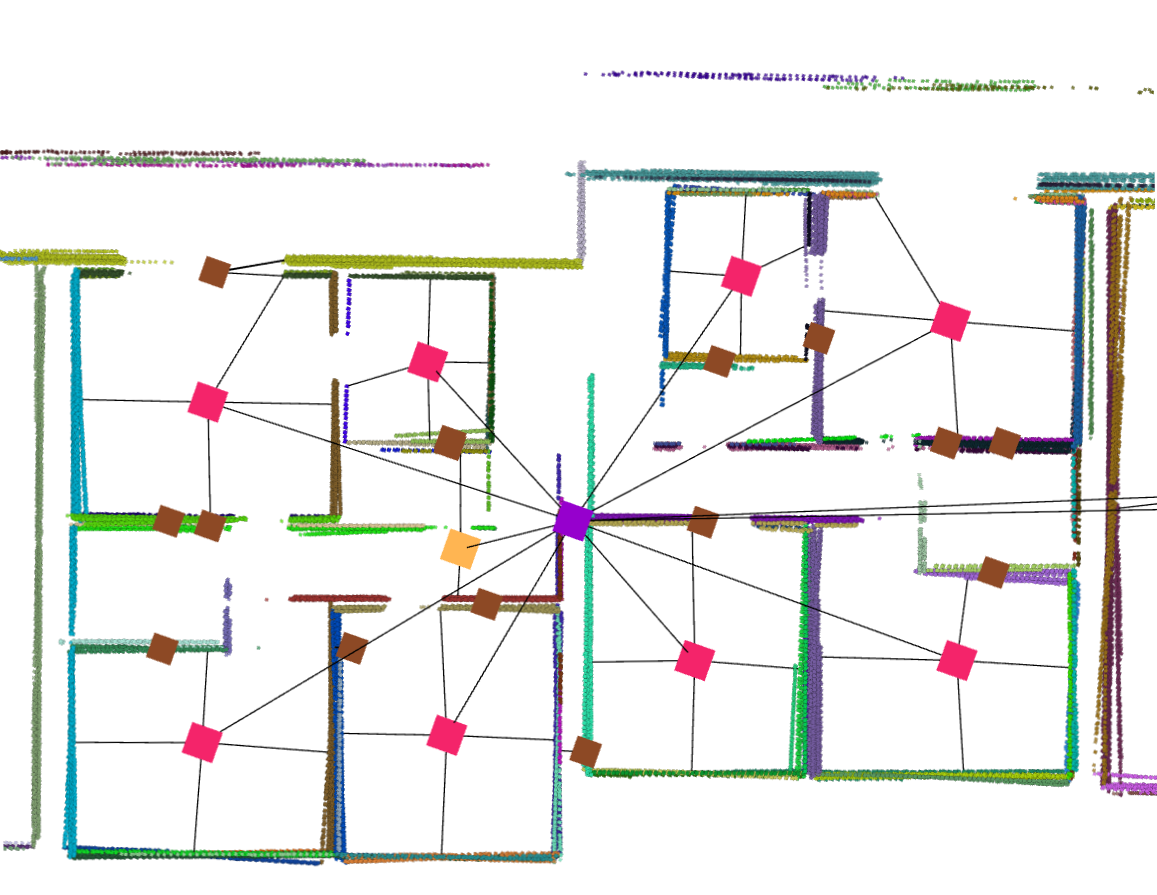} & 
    \includegraphics[height=\sh]{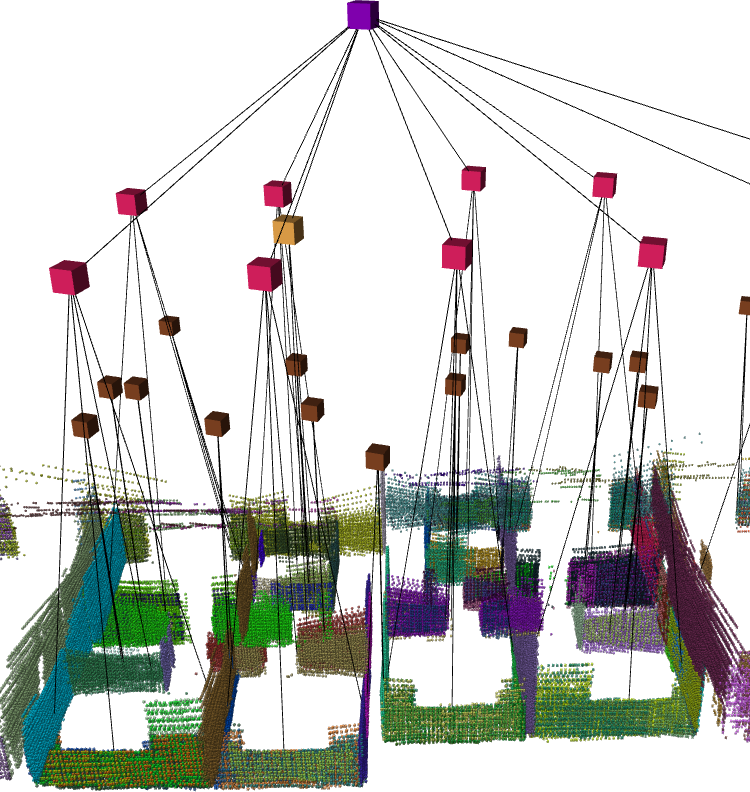} \\
    & (e) P: -/-,R: -/12 & (f) P: 7.5/9, R: 7.5/12 & (g) P: 7.5/9, R: 8/12 & (h) Walls: 14/14 \\
  \end{tabular}
  \caption{\textbf{Graph Expressiveness.} Compared qualitatively over \textit{Hydra RS}~\cite{hydra}, \textit{S-Graphs+ RS}~\cite{s_graphs+} and \textit{Ours C} on two example datasets, SE3 simulated and RC2F2 real. \textit{Room} generation is presented in the first three columns while indicating precision (P) and recall (R) ratios. The fourth column, presents a 3D perspective to demonstrate the \textit{Rooms} as well as \textit{Wall} generation but \textit{Ours C}.}
  \label{fig:rviz_examples}
\end{figure*}

\boldparagraph{Average Trajectory Error.}
The ATE for the simulated experiments is presented in Tab.~\ref{tab:ate_simulated_data}. \textit{Ours (Int.)} approach for \textit{Room} and \textit{Wall} detection demonstrates an improvement of $6.8\%$ with respect to \textit{S-Graphs+}\cite{s_graphs+} baseline. The ablation of \textit{walls} in \textit{Ours (Int.) (Rooms only)} shows an improvement of $5.2\%$ even though no \textit{Wall} entities are leveraged. Note that in simulated data resembling to real construction sites i.e., \textit{SC1F1}, \textit{SC1F2} and \textit{SE3}, our approach can of robustly detect \textit{Rooms} improving the final ATE. However, when complexity and size decrease, \textit{S-Graphs+} presents a similar or better performance due to the same number of detected rooms. 

\begin{table}[t]
\setlength{\tabcolsep}{3.8pt}
\footnotesize
\centering
\caption{\textbf{Absolute Trajectory Error (ATE)} [m], of \textit{S-Graph+} with different detection modules on simulated data. The best results are boldfaced. Both of our approaches improve the ATE of the baseline in the most complex scenes Considering semantic relations between both walls and room is always better or equal than only rooms.}
\renewcommand{\arraystretch}{1.2}
\begin{tabular}{l | l l l l l | c}
\toprule
 & \multicolumn{5}{l}{{\textbf{Dataset} [m $\times 10^{-2}$]}}  \\
\toprule
\textbf{Module} & \textit{SC1F1} & \textit{SC1F2} & \textit{SE1} & \textit{SE2} & \textit{SE3} & Avg \\
\midrule
\textit{S-Graphs+}~\cite{s_graphs+} (baseline)  & 2.72 & 6.93 & \textbf{1.47} & \textbf{1.36} &  2.98 &  3.09\\ 
\textit{Ours (Int.) (rooms only)} & 2.72 & 6.58 & 1.55 &  1.57 &  2.23 &  2.93\\
\textit{Ours (Int.)} & \textbf{2.71} & \textbf{6.35} & 1.54 &  1.56 & \textbf{2.23}  & \textbf{2.88} \\ 
\bottomrule
\end{tabular}
\label{tab:ate_simulated_data}
\end{table}
%


\boldparagraph{Map Matching Accuracy.}
Tab.~\ref{tab:rmse_real_data} presents the MMA for simulated and real experiments. \textit{Our (Int.)} presents an average improvement of $1.8\%$ with respect to \textit{S-Graphs+} \cite{s_graphs+} baseline. The ablation of \textit{Walls} in \textit{Ours (Int.) (Rooms only)} still represents an improvement of $0.3\%$. The results demonstrate that, even with an already low MMA in the baseline, the inclusion of better representations in \textit{Our (Int.)} still presents a notable improvement while enhancing its expressiveness. 

\begin{table}[tb]
\caption{\textbf{Map Matching Accuracy (MMA)} [m] of \textit{S-Graphs+} with different detection modules on simulated and real data. The best results are boldfaced. In all real scenes and all but one simulated, the MMA is outperformed by our approach, including the ablated method.
}
\centering
\scriptsize
\setlength{\tabcolsep}{1.5pt}
\renewcommand{\arraystretch}{1.2}
\begin{tabular}{l | l l l | l l l l | c}
\toprule
 & \multicolumn{8}{l}{{\textbf{Dataset} [m $\times 10^{-2}$]}}  \\
\toprule
\textbf{Module} & \scriptsize\textit{SE2} & \scriptsize\textit{SC1F1} & \scriptsize\textit{SC1F2} & \scriptsize\textit{RC1F1} & \scriptsize\scriptsize\textit{RC1F2} & \scriptsize\textit{RC2F2}  & \scriptsize\textit{RC3F2} & Avg \\
\midrule
\textit{S-Graphs+}~\cite{s_graphs+} & 27.53 & \textbf{7.40} &  7.55 & 32.60 & 18.75 & 17.8 & 44.86 & 22.35\\
\textit{Ours (Int.) (rooms only)} &  27.52 &  7.61 & 7.53 &  \textbf{32.54} & 18.64 & 17.8 & 44.35  & 22.28\\ 
\textit{Ours (Int.)} &  \textbf{27.51} &  7.60 & \textbf{7.51}  & 32.67 & \textbf{17.79} & \textbf{17.27} & \textbf{43.31} & \textbf{21.95} \\
\bottomrule 
\end{tabular}
\label{tab:rmse_real_data}
\end{table}

\boldparagraph{Limitations.}
%
The number of higher-level entities in the synthetic dataset is limited to \textit{Rooms} and \textit{Walls}, which limits the graph expressiveness of the generation. In addition, the layouts only contain a limited variety of shapes, which leads to a lower  performance on edge cases in real scenarios. Furthermore, edge postprocessing in Sec.~\ref{sec:methodology_subgraph_generation} requires the maximum cluster size for each entity of 1 for \textit{Walls} and any for \textit{Rooms} to be set in advance.
\section{Conclusion}

We presented a novel approach based on Graph Neural Networks for inferring high-level semantic-relational concepts such as \textit{Rooms} and \textit{Walls} to enrich the 3D scene graph for a given environment. 
Our method unfolds in several steps: (a) GNN-based Edge Inference: Initially, we infer \textit{``same Room"} and \textit{``same Wall"} edges among the observed low-level \textit{Planes}. (b) Clustering: Subsequently, we process these inferred edges to cluster \textit{Planes} corresponding to each higher-level concept. (c) Subgraph Generation: Finally, we represent these clusters in the form of a subgraph to form new \textit{Room} and \textit{Wall} nodes, finally incorporating them into the existing factor graph within the \textit{S-Graphs+} \cite{s_graphs+} framework.

In comparison to the current baselines for \textit{Room} segmentation, our approach exhibits a notable reduction of $67\%$ of detection time, expressiveness, and generalization attributes given the fact that \textit{Walls} entities are not yet automatically detected. Importantly these enhancements contribute to a better final pose ($6.8\%$) and map accuracy ($1.8\%$). 


In future research, we expect to expand the expressiveness of our dataset with new entities useful for the SLAM such as \textit{Floors} and more complex scenes. Furthermore, we envision the GNN-based generation of entities (\textit{Rooms} and \textit{Walls}) and their relationships in an end-to-end manner without the need of a postprocessing step.

\balance
\bibliographystyle{IEEEtran}
\bibliography{Biobliography}

\end{document}